\documentclass{article}

  \usepackage[dblblindworkshop, final]{neurips_2026}
\workshoptitle{NeurIPS 2026 Workshop TS-LIMITS}

\usepackage[utf8]{inputenc} 
\usepackage[T1]{fontenc}    
\usepackage{hyperref}       
\usepackage{url}            
\usepackage{booktabs}       
\usepackage{amsfonts}       
\usepackage{nicefrac}       
\usepackage{microtype}      
\usepackage{xcolor}         
\usepackage{graphicx}
\usepackage{subcaption}       
\usepackage{makecell}
\usepackage{amsmath} 
\usepackage{multirow} 
\usepackage{wrapfig}
\title{Sustainability of TSFM Deployment and Pretraining: From Carbon Footprint Analysis to Mitigation via Pruning}

\author{%
  Cécile Luc \\
  Strategy \& Innovation, AI Hub\\
  Schneider Electric, Rueil-Malmaison, France \\
  \texttt{cecile.luc92@gmail.com} \\
  \And
  Nathanaël Lemonnier \\
  Strategy \& Innovation, AI Hub\\
  Schneider Electric, Grenoble, France \\
  \texttt{nathanael.lemonnier@se.com} \\
  \And
  Ievgen Redko \\
  Strategy \& Innovation, AI Hub\\
  Schneider Electric, Rueil-Malmaison, France \\
  \texttt{ievgen.redko@se.com} \\
}

\begin{document}

\maketitle

\begin{abstract}
Time Series Foundation Models (TSFMs) offer strong zero-shot forecasting 
capabilities, but their high pretraining cost raises a practical 
sustainability question: does this upfront investment pay off over a full 
deployment lifecycle, compared to supervised standalone models that must be periodically 
retrained? We model the cumulative carbon cost of different approaches across 
a realistic industrial scenario: Chronos-2 
amortizes its pretraining cost within months for large or 
frequently queried deployments, but a classical model remains cheaper 
indefinitely for small or rarely queried ones. For cases where reducing 
the TSFM's own cost matters, we show that removing a single encoder block 
(7.9\% of parameters) via a gradient-free pruning criterion 
yields a consistent 6 to 9\% reduction in FLOPs, CO$_2$ emissions, and 
latency, at essentially no cost to zero-shot forecasting performance 
across a 25-task benchmark.
\end{abstract}

\section{Introduction}
\label{sec:intro}

Time Series Foundation Models (TSFMs) such as Chronos-2~\citep{ansari2025},
TimesFM~\citep{das2024timesfm}, and Moirai~\citep{woo2024moirai} have shown strong zero-shot generalization capabilities across a range of tasks, including forecasting, anomaly detection, and classification. This generalization capability comes at a price that is rarely quantified: pretraining a TSFM on massive datasets requires substantial computing resources, with an associated energy and carbon cost, paid once, upfront, unlike task-specific deep learning or classical models (e.g., PatchTST~\citep{nie2023patchtst} or LightGBM~\citep{ke2017lightgbm}), which must be periodically retrained, per asset, to follow the evolution of the data.

Carbon accounting for machine learning has received growing attention, from early estimates of the energy footprint of NLP model training (\citep{strubell2019energy}), to broader calls for treating computational efficiency as a first-class evaluation criterion alongside accuracy
(\citep{schwartz2020green}). However, to our knowledge, no existing study compares the cumulative carbon cost of TSFMs against specialized models across a full deployment lifecycle: pretraining, fine-tuning or periodic retraining, and inference. If a TSFM's upfront cost
cannot be amortized over time, it can instead be reduced directly. A common approach is pruning, which removes blocks of a pretrained model to reduce computational and memory costs with limited loss in performance.

\paragraph{Contributions.}
\begin{enumerate}
    \item \textbf{A lifecycle carbon cost comparison} of TSFMs against two representative alternatives: LightGBM, a classical gradient-boosted tree model, and PatchTST, a task-specific deep learning forecaster to determine whether a 
    TSFM's pretraining cost is amortized over a realistic deployment horizon. 
    \item \textbf{A structured pruning study of Chronos-2}, showing that 
    removing a single encoder block via a one-shot cosine similarity 
    criterion reduces inference cost by 6 to 9\% with no measurable loss 
    in zero-shot forecasting performance.
    \item \textbf{A multi-dimensional cost characterization} of both the 
    full and pruned models in terms of FLOPs, CO$_2$ emissions, and 
    latency.
\end{enumerate}

\section{Background and Related Work}

\paragraph{Time Series Foundation Model.}

TSFM architectures are highly diverse, and these differences directly affect their computational cost. Our study includes Chronos-2, an encoder-only transformer, built on the T5 design, with a group attention layer that aggregates information across time series within a batch, enabling zero-shot, multivariate, in-context forecasts without any task-specific training. This comes at a cost: self-attention scales quadratically with context length, on top of the group attention overhead. Chronos-2 does, however, produce the full forecast horizon in a single forward pass (for prediction lengths of up
to 1024), unlike earlier autoregressive 
TSFMs such as TimesFM in its first version~\citep{das2024timesfm}, which 
generated forecasts patch by patch in several steps. Regarding the size, Chronos-2 has around 120M parameters. Other recent TSFMs are much smaller: FlowState, for instance, uses under 10M parameters by replacing the transformer with a state-space encoder~\citep{graf2026flowstate}. Naturally, the cost numbers we report for Chronos-2 cannot represent TSFMs universally, but they represent our baseline for this paper.

\paragraph{Carbon Footprint and Sustainability.}
We use CodeCarbon ~\citep{schmidt2021codecarbon} to quantify the environmental cost of each approach, available as a Python package that estimates
CO$_2$-equivalent (CO$_2$eq) emissions from the energy consumed by the
underlying hardware, converted using the carbon intensity of the local
electricity grid. A common alternative is to count FLOPs, the number of arithmetic operations a model performs. Unlike carbon cost, this figure does not depend on the hardware or software stack used, which makes it a useful way to compare architectures independently of any specific measurement setup.

\paragraph{Model Compression for Time Series.}
\label{sec:related_compression}

Pruning is a common method to reduce computation and memory footprint of a model after training, both
directly relevant for the sustainability focus of our study.
Pruning removes parts of a network, weights, neurons, channels, or
entire layers, to reduce it. We distinguish two different types of pruning. Unstructured pruning removes individual weights, but the resulting sparse matrices need specialized hardware or software to actually run faster. Structured pruning removes whole components instead, such as attention heads or channels, which speeds things up on regular hardware without any extra
tooling (see \citep{zhu2024survey}). It has been studied for LLMs, but far less so for TSFMs. The study from \citet{zhao2025less} is, to our knowledge, the closest prior work
applying pruning to a TSFM. While the motivation of this paper is different, the authors prune neurons before fine-tuning, specifically to improve
downstream task performance, based on the idea that a given forecasting
task only needs a fraction of a TSFM's parameters. Our goal is not to
improve task-specific performance through pruning, but to reduce cost
while preserving the model's original zero-shot forecasting ability,
without any task-specific fine-tuning.

\section{Carbon cost of TSFM Deployment}

\subsection{Deployment scenario and cost model}
\label{sec:cost_model}

\paragraph{Problem setup.} We model a production forecasting system deployed hourly across $N_a$ industrial assets, for instance UPS units in a data center or
electrical transformers. We consider a baseline scenario: a 24-hour forecast is requested for each asset every day, and models are periodically retrained or fine-tuned to keep up with data evolution. The full notation used in this section is given in the appendix.

We consider two supervised methods as alternatives to a TSFM. The first one is LightGBM~\citep{ke2017lightgbm}, a classical gradient-boosted tree model (300 estimators in our configuration ; a fixed budget chosen without
hyperparameter tuning, see Appendix~\ref{sec:limitations}), called as a rolling forecast hour by hour, each call receiving a window of $L=168$ real sensor values, so a 24-hour forecast costs $H=24$ times a single call. We suppose it is a realistic use of a classical, lightweight model like LightGBM. The second is PatchTST~\citep{nie2023patchtst}, a patch-based transformer for 
time series forecasting (around 620K parameters in our configuration), which, like Chronos-2, is natively multi-step and produces the full 24-hour forecast in a single call.

\paragraph{Cost model.} We now introduce our cost model for all models considered above. LightGBM's cumulative
carbon cost, as a function of time $t$ (months since deployment) and
number of assets $N_a$, is:
\begin{equation}
\text{ML}(t, N_a)
= \sum_{j=1}^{N_a} \sum_{i=0}^{k_{\text{tr}}} \alpha \, W_i^{(j)}
  + \lfloor t \rfloor \; N_a \; n_{\text{inf}} \; H \; C_{\text{inf}}^{\text{C}}
\end{equation}
with $k_{\text{tr}} = \lfloor (t-1) / T_{\text{tr}} \rfloor$ for $t \geq 1$,
and $\text{ML}(0, N_a) = 0$. The double sum adds up the cost
of every retraining that has happened so far: one LightGBM model is
trained from scratch per asset, then retrained every $T_{\text{tr}}$
months, and each of these retrainings costs $\alpha$ (the carbon cost of
training on a single window) times the size of the retraining window at
that point, $W_i^{(j)}$. The remaining term is the inference cost:
$n_{\text{inf}}$ is the number of 24-hour forecasts requested per month
and per asset, and $C_{\text{inf}}^{\text{C}}$ is the cost of a single
LightGBM call, so that each forecast costs $H \times C_{\text{inf}}^{\text{C}}$, as described above.

Chronos-2 using zero-shot, by contrast, is never retrained:
\begin{equation}
\text{TSFM}_{\text{ZS}}(t, N_a)
= C_{\text{pr}}
  + \lfloor t \rfloor \; N_a \; n_{\text{inf}} \; C_{\text{inf}}^{\text{T}}
\end{equation}
$C_{\text{pr}}$, the cost of pretraining, is paid once and does not
depend on $N_a$, and $C_{\text{inf}}^{\text{T}}$ is the cost of a single
Chronos-2 call, which already covers the full 24-hour forecast. PatchTST
follows the same logic as LightGBM but without the $H$ factor, since it
is also natively multi-step, and Chronos-2 with LoRA or full
fine-tuning follows the same logic as the zero-shot case with an added
periodic fine-tuning term. We give the full equations, parameter notation (Table~\ref{tab:notation}), and their measured values (Table~\ref{tab:parameter_values}) in the appendix (Section~\ref{sec:appendix_equations}).


\paragraph{Measurement protocol.}
\label{sec:protocol}
Carbon is measured with CodeCarbon~\citep{schmidt2021codecarbon} at the local grid carbon intensity 56 gCO$_2$/kWh, to which all values are \textbf{proportional}. Chronos-2's undisclosed \textbf{pretraining cost} is \textbf{extrapolated from one measured step}, hence a \textbf{lower bound}. For details about those two points, see Appendix~\ref{sec:appendix_protocol}.

\subsection{Results}
\label{sec:results}
\paragraph{Raw measures.}
Table~\ref{tab:raw_measures_main} reports the measured carbon cost and 
FLOPs for each approach. We note that Chronos-2, despite being a foundation model, is one order of magnitude more carbon-intensive at inference than both supervised baselines. PatchTST is on par with LightGBM with respect to this metric as the latter provides forecasts for 1 step at a time. Finally, the full pre-training of Chronos-2 for 200,000 steps would roughly make it pre-training 60 times more carbon-intensive than PatchTST. 

\begin{wraptable}{r}{0.52\textwidth}
    \centering
    \vspace{-1em}
    \caption{Carbon cost and FLOPs per approach (LightGBM: CPU; 
    others: GPU). Full measurements and details in Appendix~\ref{sec:appendix_protocol}.}
    \label{tab:raw_measures_main}
    \footnotesize
    \resizebox{\linewidth}{!}{
    \begin{tabular}{llcc}
        \toprule
        Model & Operation & Carbon (gCO$_2$) & FLOPs \\
        \midrule
        \multirow{2}{*}{LightGBM} & Training  & $7.2 \times 10^{-3}$ & $5.3 \times 10^{10}$ \\
                                  & Inference & $6.8 \times 10^{-6}$ & $2{,}973$ \\
        \midrule
        \multirow{2}{*}{PatchTST} & Training  & $1.08$ & $5.5 \times 10^{14}$ \\
                                  & Inference & $6.6 \times 10^{-6}$ & $7.6 \times 10^{8}$ \\
        \midrule
        \multirow{2}{*}{Chronos-2} & Pretraining (1 step) & $3.3 \times 10^{-4}$ & $1.5 \times 10^{12}$ \\
                                   & Inference            & $6.1 \times 10^{-5}$ & $1.5 \times 10^{9}$ \\
        \bottomrule
    \end{tabular}}
\end{wraptable}

\begin{figure}[h]
  \centering
  \includegraphics[width=0.75\textwidth]{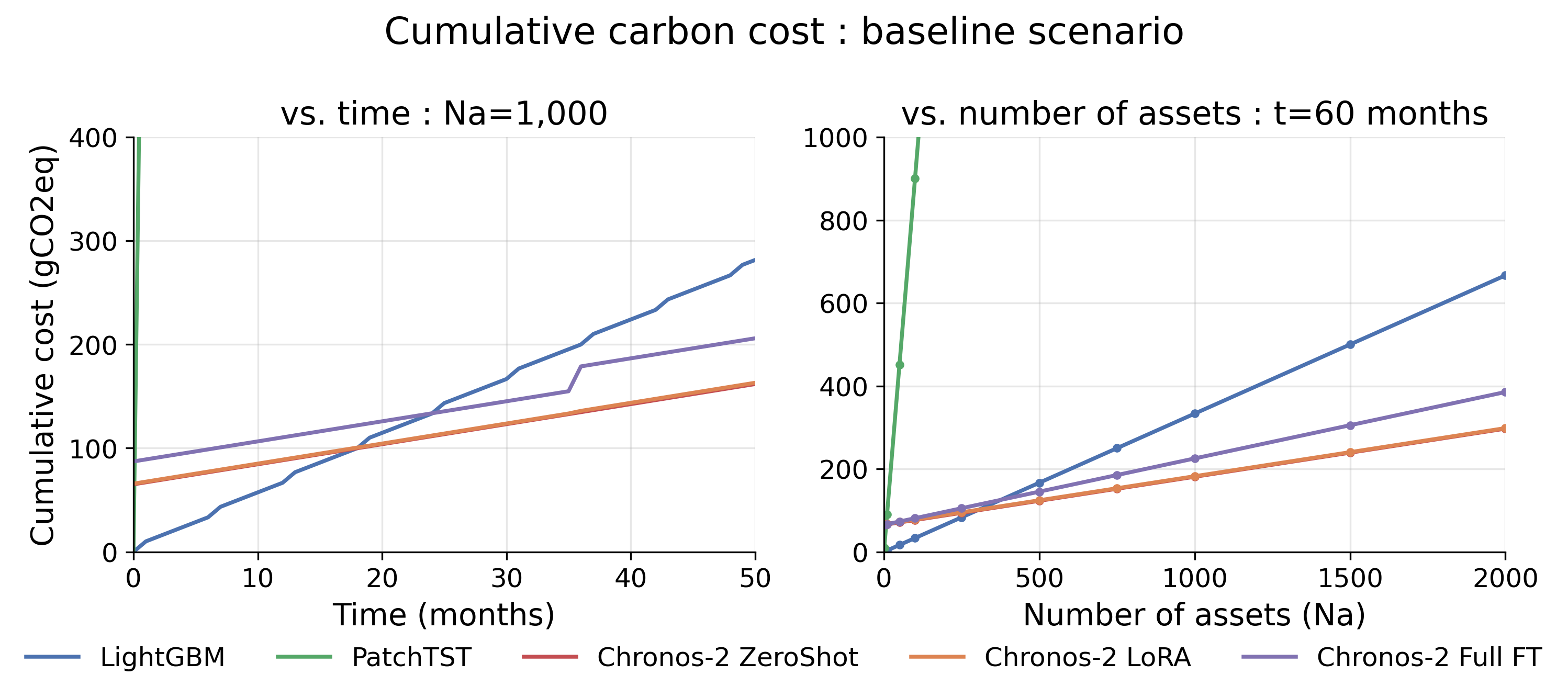}
  \caption{Cumulative carbon cost for all five approaches, baseline
  scenario, versus time at $N_a=1000$ (left) and versus number of assets at
  $t=60$ months (right).}
  \label{fig:new/fig1_carbon_baseline}
\end{figure}

\paragraph{Baseline comparison.}
Figure~\ref{fig:new/fig1_carbon_baseline} shows that, under the baseline
scenario, LightGBM starts at zero cost and grows steadily, with small
jumps every 6 months from periodic retraining, while Chronos-2 ZeroShot
starts higher, from paying $C_{\text{pr}}$ upfront, but grows more
slowly, so eventually, the two curves cross. PatchTST's training cost
per window, $\alpha_{\text{ptst}}$, is orders of magnitude higher than
LightGBM's (Table~\ref{tab:parameter_values} in Appendix). It exceeds the plotted range
within a few months, or a few dozen assets, and never becomes
competitive with LightGBM or Chronos-2 under this scenario. This assumes PatchTST is trained per asset; a PatchTST shared across
similar assets would not scale with $N_a$ in training cost
(Appendix~\ref{sec:limitations}).

Where this crossover
happens depends on both $N_a$ and $t$: at $N_a=1000$, Chronos-2 becomes
cheaper after 18 months; at $N_a=5000$, after less than 3 months;
at $N_a=100$, only after 180 months (15 years); at $N_a=10$, it
does not become cheaper even within 100 years (see
Appendix, Figure~\ref{fig:new/fig2_t_star_vs_na.png}, for the full curve).
For a small number of assets or a short deployment, a classical model stays cheaper; the TSFM only pays off once the number of assets or the usage duration is large enough to amortize $C_{\text{pr}}$.

\paragraph{Sensitivity to deployment context.} We test how sensitive this break-even point is to two deployment parameters: (i) how often LightGBM is retrained and Chronos-2 is fine-tuned
($T_{\text{tr}}$, $T_{\text{ft}}$, in months), modeling the need to
retrain more or less often as data drifts, and (ii) how often a forecast is
requested ($n_{\text{inf}}$). We consider four retraining/fine-tuning scenarios: severe drift 
($T_{\text{tr}}=1$, $T_{\text{ft}}=6$), baseline ($T_{\text{tr}}=6$, 
$T_{\text{ft}}=36$), mild drift ($T_{\text{tr}}=12$, $T_{\text{ft}}=60$), 
and a utopia scenario where neither model is ever retrained, used as an 
asymptotic reference. We also vary inference frequency across three 
scenarios: hourly ($n_{\text{inf}}=720$/month), daily ($n_{\text{inf}}=30$/month), 
and weekly ($n_{\text{inf}}\approx4$/month).

\begin{figure}[h]
    \centering
    \includegraphics[width=0.49\linewidth]{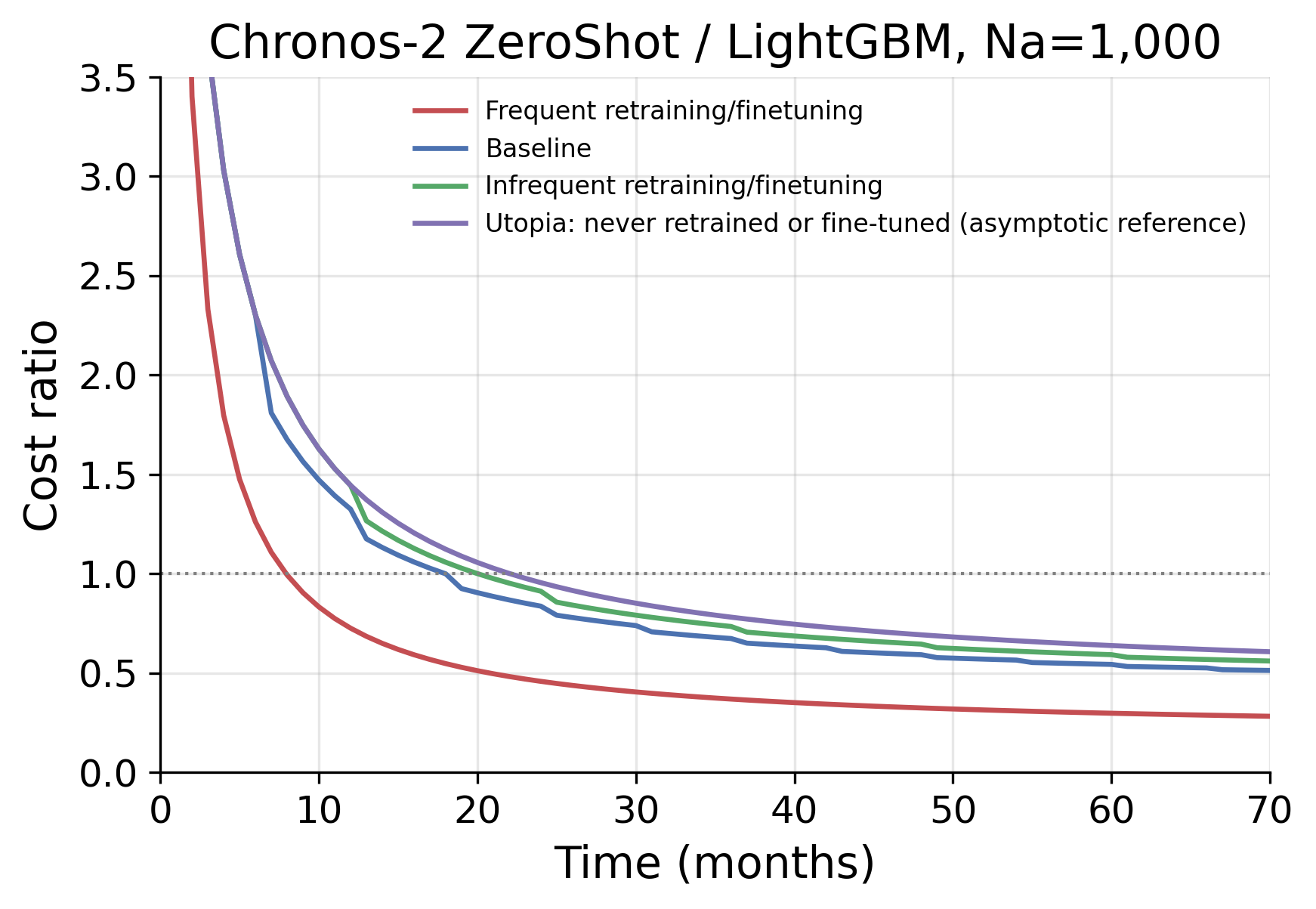}
    \hfill
    \includegraphics[width=0.49\linewidth]{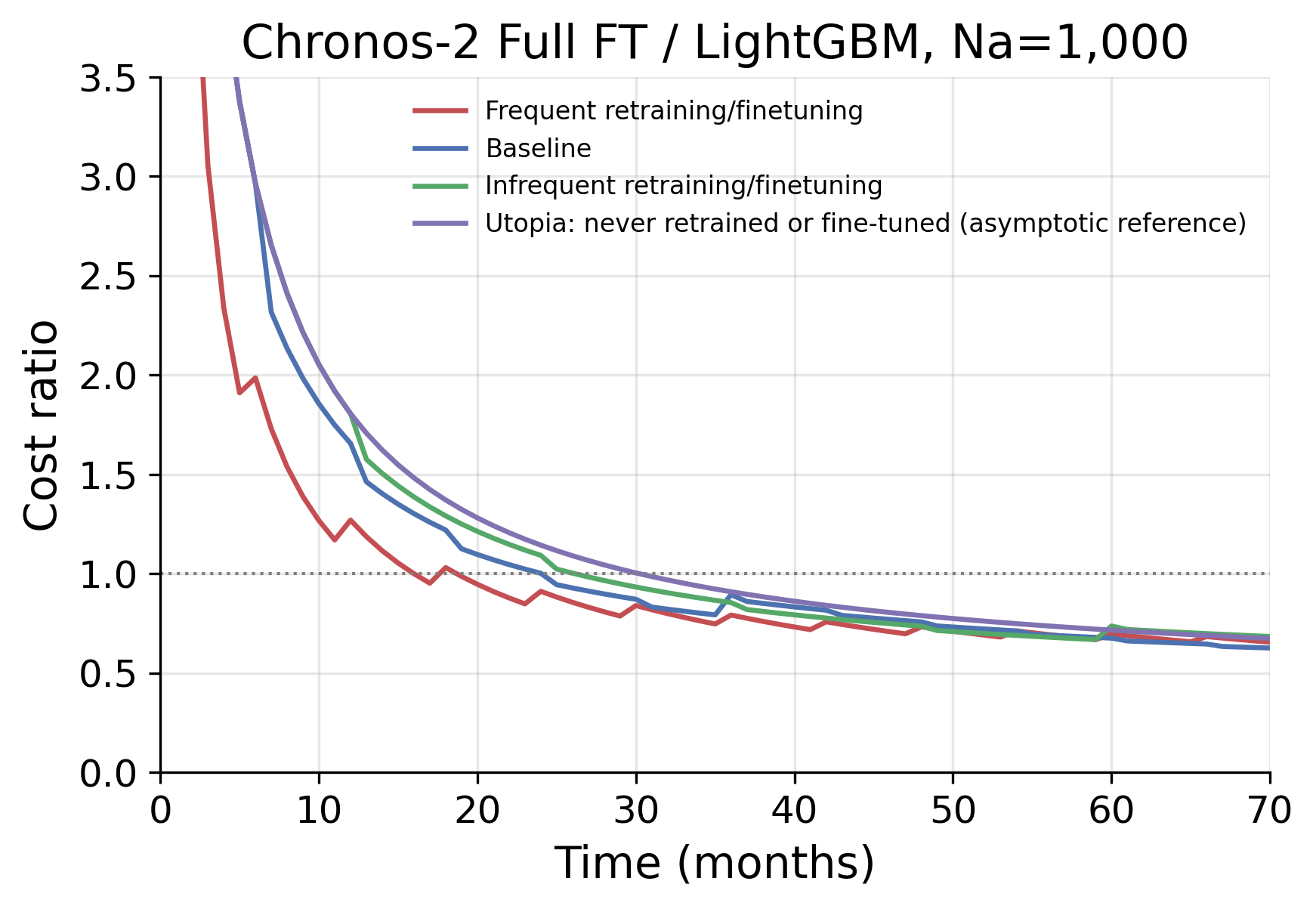}
    \caption{Cost ratio of Chronos-2 ZeroShot (left) and full 
    fine-tuning (right) over LightGBM, at $N_a=1{,}000$, under four 
    retraining/fine-tuning scenarios.}
    \label{fig:fig3_drift_sensitivity}
\end{figure}

We look at the drift scenarios under two variants of Chronos-2 (Figure~\ref{fig:fig3_drift_sensitivity}): ZeroShot,
whose cost equation has no fine-tuning term at all, and full
fine-tuning, which occurs every $T_{\text{ft}}$ months like
LightGBM is retrained every $T_{\text{tr}}$ months. This lets us separate
two effects: how much LightGBM's own retraining cost changes with
drift, and how much Chronos-2's break-even point shifts once it also
has to pay a periodic fine-tuning cost to keep up.

At $N_a=1000$, ZeroShot's break-even usage duration (where the lines crosses 1) shifts from 8 months under
severe drift, to 22 under utopia (18 in the baseline, 20 under mild
drift), purely from LightGBM's retraining cost changing. Full
fine-tuning follows the same trend, shifted later by roughly 6-8 months
in every scenario, the added cost of keeping Chronos-2 itself up to
date (Figure~\ref{fig:fig3_drift_sensitivity}).

\begin{figure}[h]
    \centering
    \vspace{1mm}
    \begin{minipage}{0.55\linewidth}
        \includegraphics[width=\linewidth]{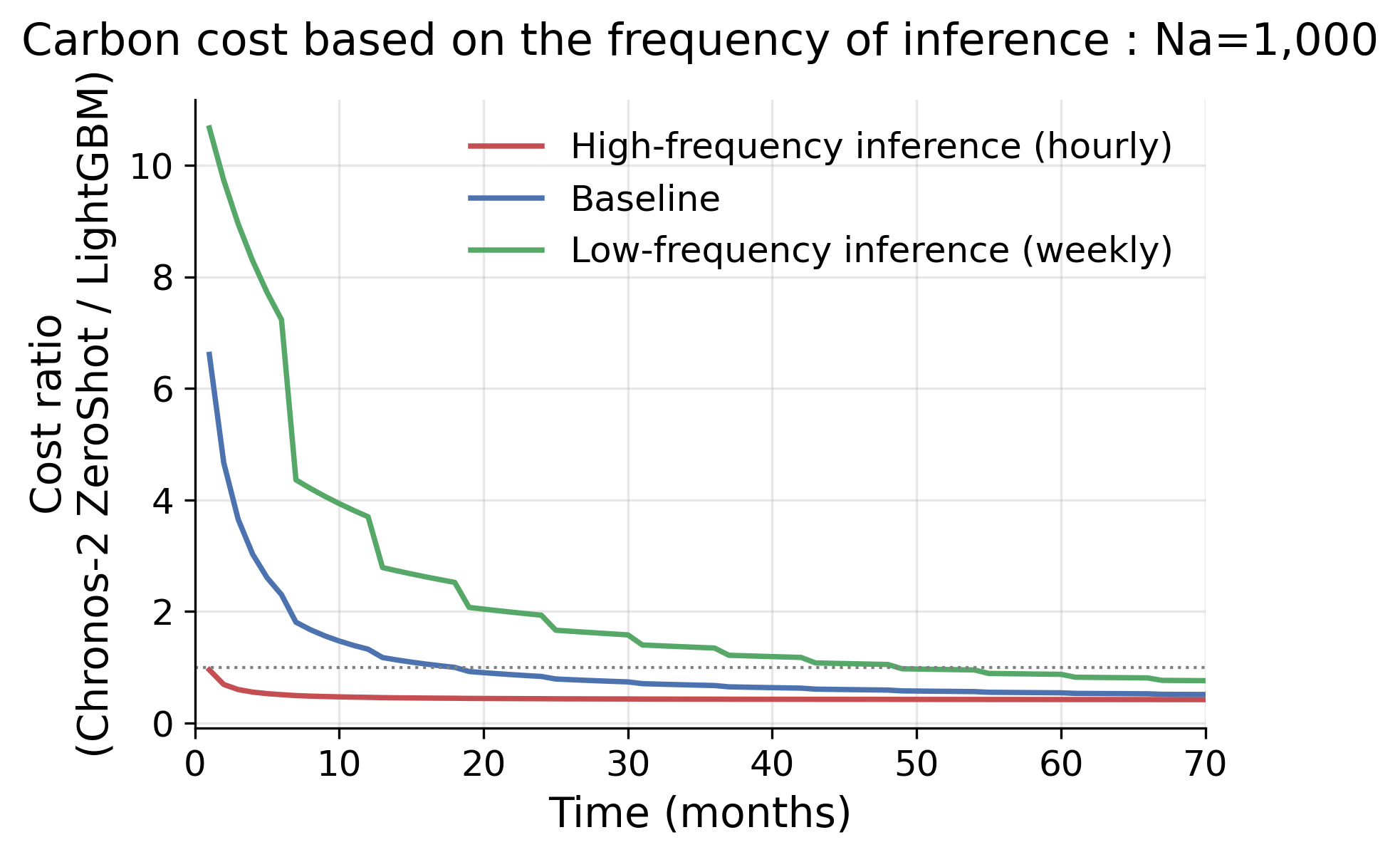}
        \caption{Cost ratio of Chronos-2 ZeroShot over LightGBM, at 
        $N_a=1{,}000$, under three inference frequency scenarios.}
        \label{fig:fig4_frequency_sensitivity}
    \end{minipage}
    \hfill
    \begin{minipage}{0.39\linewidth}
        Overall, a small or rarely queried set of assets stays cheaper with a
        classical model over any realistic usage duration, while a large or
        frequently queried one amortizes Chronos-2's pretraining cost within a
        few months. Cases in between depend on the specific usage duration,
        number of assets, and query frequency, and should be read against
        Figures~\ref{fig:new/fig1_carbon_baseline}--\ref{fig:fig4_frequency_sensitivity}.
        
        When the TSFM does not amortize its pretraining cost, reducing its inference cost directly becomes the question, which we turn to below.
    \end{minipage}
\end{figure}

Inference frequency has a much larger effect (Figure~\ref{fig:fig4_frequency_sensitivity}): Chronos-2 ZeroShot is cheaper within the first month under hourly requests, at 18 months under baseline, and only after 49 months under weekly requests. \textbf{How often a forecast is requested matters more to carbon cost than how often the model is kept up to date.}

\section{Model Compression with Pruning}

Having established Chronos-2's carbon footprint relative to classical
models, we now investigate whether that footprint can be
reduced through model compression, and at what cost to accuracy. We
focus on structured pruning: removing entire encoder blocks from the
pretrained model, without any fine-tuning, so that any efficiency gain
comes purely from a smaller architecture applied zero-shot. Concretely,
we ask: (i) can we identify redundant encoder blocks using a
lightweight, gradient-free signal; (ii) how much of the model can be
removed before accuracy degrades meaningfully; and (iii) does the
resulting model meaningfully reduce inference-time CO$_2$, latency, and
compute cost.

\subsection{Problem setup}
\label{sec:setuppruning}

\paragraph{Pruning method.}
Chronos-2's encoder is composed of 12 transformer blocks, each
alternating time attention and group attention sub-layers followed by a
feed-forward network. We use structured block-level pruning based on the
cosine similarity between the input and output of each block, following outcomes from \citet{gromov2024unreasonable}.
A score close to 1 means the block's output is nearly identical to its input (i.e., it applies a near-identity transformation) and is a strong removal candidate. The original method
targets causal decoder models and only looks at the last token. As
Chronos-2 is an encoder-only, bidirectional architecture, we instead average the cosine similarity across all context positions for a more robust per-block score (the input to each block is a patch embedding covering multiple raw time steps, hence multiple positions). Removing a block removes its time attention, group attention, and feed-forward
network (FFN) sub-layers all at once, so no single mechanism can be isolated at this granularity. This scoring gives a redundancy ranking of the 12 blocks. Given a
compression budget of $k$ blocks, we remove the $k$ highest-ranked
(most redundant) blocks as a group; the ranking is computed once, on a
dedicated calibration set, and reused for every $k$ we test.

\paragraph{Dataset.}
We evaluate using 27
zero-shot forecasting tasks spanning diverse domains and frequencies
(30~min to yearly) from \emph{Chronos Benchmark~II} \citep{ansari2024chronos}. None of the tasks were seen during Chronos-2 pretraining. Task-level
statistics are reported in Table~\ref{tab:benchmark2_stats} (Appendix).
Datasets are downloaded from HuggingFace
(\texttt{autogluon/chronos\_datasets}). Two tasks (ETT 15-min, ETT
hourly) are excluded due to a loading incompatibility (deprecated
\emph{dataset script}) with the current \texttt{datasets} library. Our evaluation covers the remaining 25 tasks.

\paragraph{Calibration and configuration selection.} \label{par:config_selection} Six tasks are used for calibration (\emph{ercot}, \emph{exchange\_rate}, \emph{monash\_hospital}, \emph{dominick}, \emph{monash\_m3\_monthly}, \emph{m4\_yearly}), covering a range of domains, frequencies, and context lengths. For each, up to 30 series are sampled by history length, with early stopping once the median cosine score stabilizes ($<$1\% change over 5 series). Scores are aggregated by median within each task, then by equal-weight average across tasks, so large tasks (\emph{dominick}, 100k+ series) do not dominate small ones (\emph{ercot}, 8 series). Using the resulting ranking, we test compression budgets $k \in \{0,\dots,6\}$ on these same 6 tasks and keep the $k$ that maximizes the aggregated skill score. Final evaluation adds the 19 held-out tasks never used for calibration.

\paragraph{Evaluation protocol.}
We follow the original Chronos Benchmark~II protocol
\citep{ansari2024chronos}: a single, predefined test window per series (task-specific horizon $H$, 4--56 steps), with all preceding history used as context. All evaluation runs in strict univariate mode (\texttt{cross\_learning=False}), isolating the effect of pruning from in-context learning across series.

\paragraph{Metrics.} We report MASE for point-forecast accuracy and WQL for probabilistic quality, using standard GluonTS conventions; MASE seasonality is inferred automatically from each series' frequency (\texttt{get\_seasonality}), following \citet{ansari2024chronos} (Appendix~C). Scores are aggregated across tasks using a geometric mean of the ratio between the model's metric and a Seasonal Naive baseline, with equal weight per task, following the aggregation convention of \citet{ansari2024chronos}. We report this aggregate as a skill score, $100 \times (1 - \mathrm{gmean}(r))$ (higher is better), the convention used by \citet{ansari2025}. We also report a win rate: the share of tasks on which the pruned model beats the full model.

CO$_2$, latency, and FLOPs measurements details in Appendix~\ref{sec:appendix_protocol_pruning}.

\subsection{Results}
\label{sec:results_model}
\paragraph{Block redundancy and configuration selection.}

\begin{figure}[h]
    \centering
    \begin{subfigure}[t]{0.495\linewidth}
        \centering
        \includegraphics[width=\linewidth]{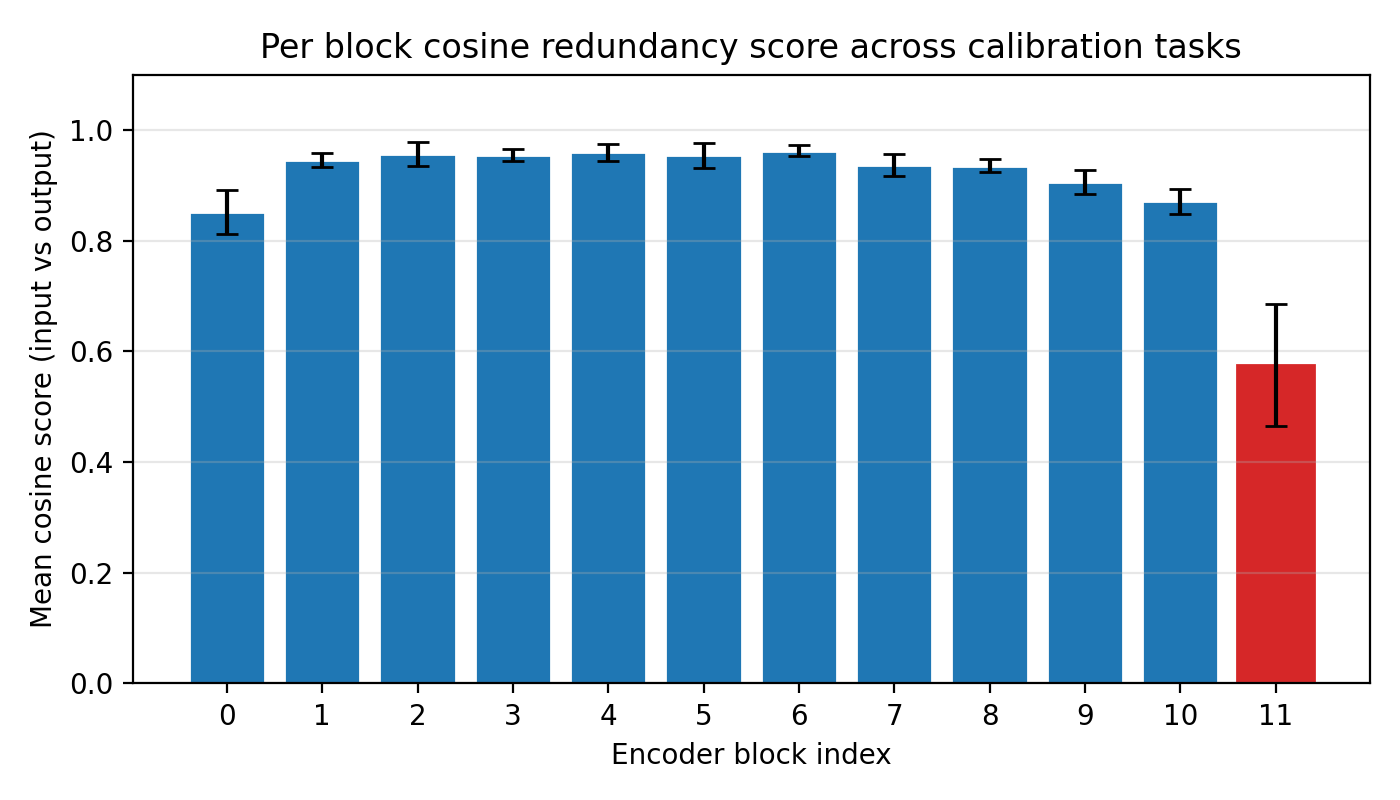}
        \caption{Cosine score per block.}
        \label{fig:cosine_scores}
    \end{subfigure}
    \hfill
    \begin{subfigure}[t]{0.495\linewidth}
        \centering
        \includegraphics[width=\linewidth]{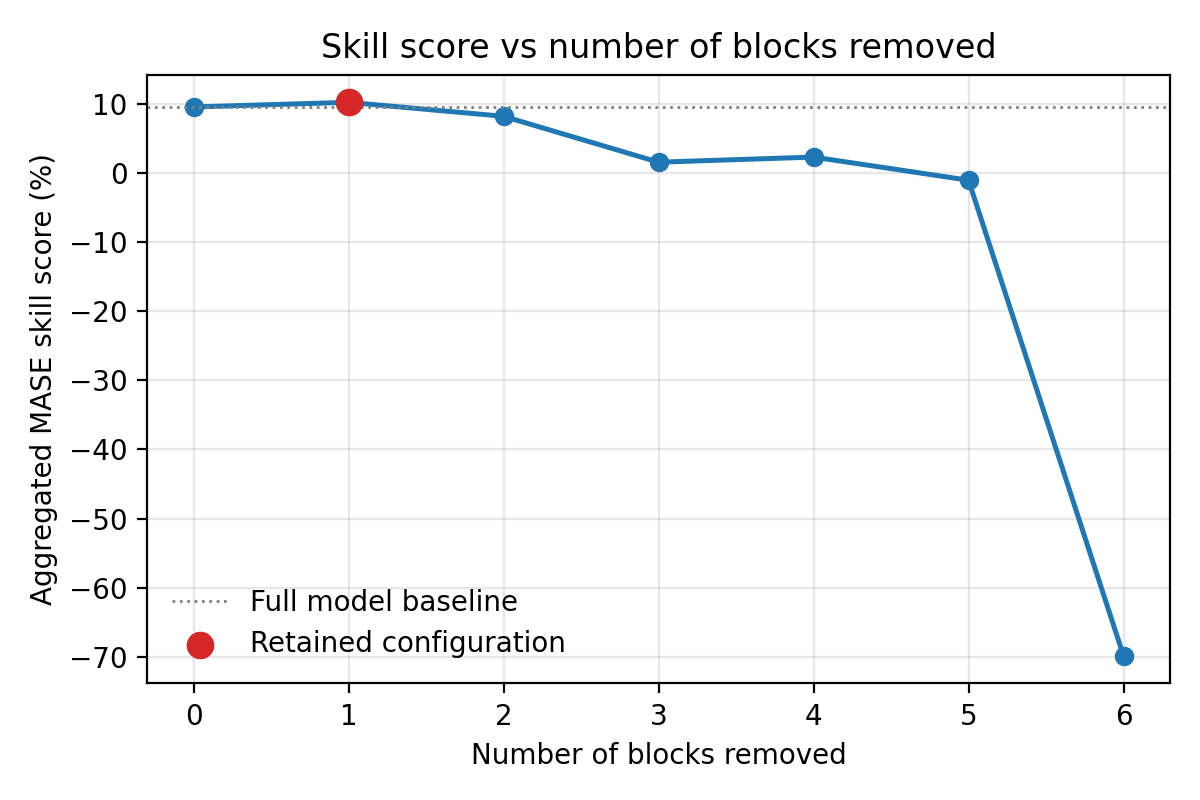}
        \caption{Skill score vs blocks removed.}
        \label{fig:skill_vs_blocks}
    \end{subfigure}
    \caption{(a) Mean cosine similarity score per Chronos-2 encoder block, averaged across the 6 calibration tasks (error bars: std across tasks). (b) Aggregated MASE skill score vs number of blocks removed on the same tasks; the dotted line marks the full model baseline, and the retained configuration (1 block) is highlighted.}
    \label{fig:pruning_calibration}
\end{figure}

Figure~\ref{fig:cosine_scores} reports
the per-block cosine score obtained during calibration. Block 11 (the last encoder block) has by far the lowest score (0.576) and the highest inter-task standard deviation (0.111). It scores lower than the median of the other 11 blocks on all 6 calibration tasks, a difference confirmed by a one-sided Wilcoxon signed-rank test ($p = 0.016$), indicating that this block is consistently the least redundant one across task domains. The remaining blocks fall in a narrower, higher range (0.85 to 0.96), with lower inter-task variance.
This is consistent with the few-layer dominance reported for large time
series models by \citet{qiu2026revealing}: only a small subset
of layers is functionally important, while the majority are redundant.

Figure~\ref{fig:skill_vs_blocks} shows the aggregated MASE skill score on
the 6 calibration tasks as blocks are progressively removed, following
the cosine similarity ordering. Removing a single block (the most
redundant one, ranked first by the cosine score) slightly improves the
skill score over the full model (10.24\% vs.\ 9.57\%). Performance then
degrades gradually up to 3 blocks removed, before dropping sharply at 6
blocks removed ($-69.9\%$), suggesting that beyond a certain point the
blocks being removed are no longer redundant but functionally important. 
Based on this exploration, we retain the single-block configuration
($k=1$) for the remainder of our study, as it is the only tested
configuration that does not degrade the aggregated skill score relative
to the full model.

\subsection{Full-scale evaluation}

We confirm this result on the complete 25-task evaluation set, including
19 tasks never seen during calibration or configuration selection.
Table~\ref{tab:full_vs_pruned} summarizes the comparison between the full
model and the retained pruned configuration (1 block removed, 7.9\%
fewer parameters).

\begin{wraptable}{r}{0.48\textwidth}  
    \centering
    \small                            
    \caption{Aggregated performance comparison between the full and
    pruned Chronos-2 models on the 25-task evaluation set. Skill scores
    are computed relative to a Seasonal Naive baseline (higher is
    better). Win rate is the share of tasks on which the pruned model
    beats the full model.}
    \label{tab:full_vs_pruned}
    \setlength{\tabcolsep}{4pt}       
    \begin{tabular}{lccc}
        \toprule
        Metric & Full & Pruned & $\Delta$ \\
        \midrule
        MASE skill score (\%)    & 22.99 & 21.87 & $-1.12$ \\
        WQL skill score (\%)     & 41.24 & 41.53 & $+0.29$ \\
        MASE win rate (\%)       & \multicolumn{2}{c}{--} & 40.0 \\
        WQL win rate (\%)        & \multicolumn{2}{c}{--} & 52.0 \\
        \bottomrule
    \end{tabular}
\end{wraptable}

The pruned model shows a small drop in point-forecast accuracy (MASE
skill score, $-1.12$ points, about 5\% relative) but a small improvement
in probabilistic forecast quality (WQL skill score, $+0.29$ points). We do not have a verified explanation for this split between MASE and
WQL, and both differences are small. A possible factor, which we did not
test, is that MASE only evaluates the median whereas WQL averages the
loss over nine quantile levels. Overall, removing the single block selected by our criterion (block 6) changes aggregated performance by roughly one skill-score point at most on both metrics, on tasks that were not used during the pruning procedure. Per-task results are reported in Table~\ref{tab:per_task_results} (Appendix).

\subsection{Compute cost reduction}

Table~\ref{tab:compute_cost} reports parameter count, FLOPs (computed analytically for a single forward pass), measured CO$_2$
emissions, and measured latency for the full and pruned models. More details in Section \ref{sec:appendix_protocol_pruning}.
\begin{table}[h]
    \centering
    \caption{Compute cost comparison between the full and pruned
    Chronos-2 models.}
    \label{tab:compute_cost}
    \begin{tabular}{lccc}
        \toprule
        Metric & Full & Pruned & Reduction \\
        \midrule
        Parameters               & 119.48M           & 110.04M           & 7.90\% \\
        FLOPs                    & 59.52\,G          & 54.67\,G          & 8.15\% \\
        CO2 emissions (g CO2eq)  & $1.221 \pm 0.037$ & $1.114 \pm 0.002$ & 8.72\% \\
        Latency (ms)              & $19.43 \pm 1.02$  & $18.27 \pm 1.00$  & 5.97\% \\
        \bottomrule
    \end{tabular}
\end{table}

All four independent measures of computational cost, parameter count,
analytically-computed FLOPs, measured CO$_2$ emissions, and measured
latency, show a consistent reduction in the 6 to 9\% range, closely
matching the 7.9\% parameter reduction achieved by removing a single
encoder block. This agreement across independent metrics reinforces the
reliability of our measurements. Latency shows the smallest relative
reduction, which is expected since it includes fixed overheads
(data loading, tokenization, CUDA synchronization) that do not scale down
with the removed block.

Taken together, these results show that a small, targeted pruning of
Chronos-2 (the single block selected by our criterion) reduces inference-time
compute cost and CO$_2$ emissions by roughly 6 to 9\%, at essentially no
cost to aggregated forecasting performance on a 25-task zero-shot
benchmark.

\section{Conclusion}
\label{sec:conclusion}
This paper examined the sustainability of TSFM deployment from two angles. First, we showed that Chronos-2's pretraining cost is only amortized under sufficiently large or frequently queried deployments: at $N_a=1{,}000$ assets, it breaks even with LightGBM after about 18 months under daily inference, but this threshold shifts by up to $49\times$ with inference frequency alone, and a classical model remains cheaper indefinitely for a small number of assets. Second, removing a single encoder block (7.9\% of parameters) selected by a gradient-free pruning criterion yields a consistent 6 to 9\% reduction in FLOPs, CO$_2$ emissions, and latency, at essentially no cost to zero-shot forecasting performance. Together, these results offer a practical framework for sustainable TSFM deployment, and a low-effort compression path when pretraining cost does not amortize. See limitations of our study in Appendix~\ref{sec:limitations}.

\clearpage

\bibliographystyle{plainnat}
\bibliography{cecile_tsfm_tslimits}

\clearpage

\appendix

\section{Appendix 1: Cost Model Details}
\label{sec:appendix_equations}

\subsection{Notation}

\begin{table}[h]
  \centering
  \caption{Notation used in the cost equations.}
  \label{tab:notation}
  \begin{tabular}{lll}
    \toprule
    Symbol & Unit & Description \\
    \midrule
    $t$ & months & Time elapsed since deployment. \\
    $N_a$ & -- & Number of assets. \\
    $n_{\text{inf}}$ & inf./month & Number of 24h forecasts requested per month, per asset. \\
    $W_i$ & windows & Size of the training set used at the $i$-th retraining. \\
    $a, b$ & windows & Bounds of $W_i \sim \mathcal{U}[a,b]$. \\
    $\alpha$ & gCO$_2$/window & Carbon cost of training LightGBM on a single window. \\
    $T_{\text{tr}}$ & months & Time between two LightGBM/PatchTST retrainings. \\
    $H$ & -- & Number of successive calls per 24h forecast for LightGBM ($H=24$). \\
    $C_{\text{inf}}^{\text{C}}$ & gCO$_2$ & Carbon cost of a single LightGBM inference call. \\
    $\alpha_{\text{ptst}}$ & gCO$_2$/window & Carbon cost of training PatchTST on a single window. \\
    $C_{\text{inf}}^{\text{ptst}}$ & gCO$_2$ & Carbon cost of a single PatchTST inference call. \\
    $C_{\text{pr}}$ & gCO$_2$ & Carbon cost of pretraining Chronos-2, paid once. \\
    $C_{\text{inf}}^{\text{T}}$ & gCO$_2$ & Carbon cost of a single Chronos-2 inference call. \\
    $C_{\text{ft}}^{\text{LoRA}}$ & gCO$_2$ & Carbon cost of one Chronos-2 LoRA fine-tuning run. \\
    $T_{\text{ft}}$ & months & Time between two Chronos-2 fine-tunings (LoRA or full). \\
    $\beta$ & gCO$_2$/window & Carbon cost of full fine-tuning Chronos-2 on a single window. \\
    $p$ & -- & Fraction of each asset's data included in the full fine-tuning sample. \\
    $W_{\text{Full}}(N_a)$ & windows & Resulting size of the full fine-tuning sample: $p \times W_{\text{base}} \times N_a$. \\
    \bottomrule
  \end{tabular}
\end{table}

\subsection{Full cost equations}

PatchTST follows the same logic as LightGBM but without the $H$ factor,
since it is natively multi-step:
\begin{equation}
\text{PatchTST}(t, N_a)
= \sum_{j=1}^{N_a} \sum_{i=0}^{k_{\text{tr}}} \alpha_{\text{ptst}} \, W_i^{(j)}
  + \lfloor t \rfloor \; N_a \; n_{\text{inf}} \; C_{\text{inf}}^{\text{ptst}}
\end{equation}

Chronos-2 with LoRA fine-tuning adds a periodic fine-tuning term to the
zero-shot cost:
\begin{equation}
\text{TSFM}_{\text{LoRA}}(t, N_a)
= C_{\text{pr}}
  + (k_{\text{ft}} + 1) \; C_{\text{ft}}^{\text{LoRA}}
  + \lfloor t \rfloor \; N_a \; n_{\text{inf}} \; C_{\text{inf}}^{\text{T}}
\end{equation}
with $k_{\text{ft}} = \lfloor t / T_{\text{ft}} \rfloor$. We model a
single LoRA adapter trained on a representative sample of a few assets
and shared across the assets, so its cost does not depend on $N_a$. A
first fine-tuning is performed at $t=0$, then every $T_{\text{ft}}$
months. We suppose it is a realistic way to fine-tune with LoRA.

Chronos-2 with full fine-tuning follows the same structure, but the
fine-tuning cost now depends on $N_a$:
\begin{equation}
\text{TSFM}_{\text{Full}}(t, N_a)
= C_{\text{pr}}
  + \sum_{j=0}^{k_{\text{ft}}} \beta \, W_{\text{Full}}(N_a)
  + \lfloor t \rfloor \; N_a \; n_{\text{inf}} \; C_{\text{inf}}^{\text{T}}
\end{equation}
with $W_{\text{Full}}(N_a) = p \times W_{\text{base}} \times N_a$, and
$W_{\text{base}} = (a+b)/2$: full fine-tuning is run on a stratified
sample of a fixed fraction $p$ of each asset's data, so the fine-tuning
sample size grows linearly with the number of assets. We suppose it is a realistic way to full fine-tune.

\section{Appendix 2: Measurement Protocol Details (cost model)}
\label{sec:appendix_protocol}

\paragraph{Hardware and carbon intensity.}
All measurements use an NVIDIA A100 80GB PCIe GPU (Azure
\texttt{Standard\_NC24ads\_A100\_v4} instance, \texttt{francecentral}
region), paired with an AMD EPYC 7V13 CPU. CodeCarbon (v3.2.9) reads GPU
power directly through \texttt{nvidia-ml-py}, while CPU power is
estimated from a TDP-based model, since the RAPL hardware counters it
otherwise reads directly were not accessible on our virtual machine.
This estimation method is applied identically across all measurements. The carbon intensity CodeCarbon applied
throughout, computed directly from the emissions and energy figures
logged for each run, is $\approx$56 gCO$_2$/kWh for Île-de-France,
France, consistent with historical French grid intensity figures from
Our World in Data (e.g., 56.54 gCO$_2$/kWh in July 2020).

CodeCarbon applies the electricity carbon intensity of the detected
country, taken from Our World in Data, or of the cloud provider region
when available.\textbf{ Every carbon value is an energy times the local grid carbon intensity
applied by CodeCarbon, so it is proportional to it (multiply by $I/56$
for a grid of $I$ gCO$_2$/kWh). If pretraining and deployment run on the
same grid, all costs scale together and the break-even durations are
unchanged.}


\paragraph{Estimating $C_{\text{pr}}$.}
Chronos-2 is never pretrained from scratch in this work. We estimate
$C_{\text{pr}}$, the fixed cost a TSFM deployment must amortize and the
parameter driving every break-even result in Section~\ref{sec:results_model},
by measuring a single pretraining step and extrapolating from it: one
forward and backward pass on a heterogeneous batch combining univariate
series, multivariate groups, and series with covariates, with a context
length of 2048, following the first stage of Chronos-2's own
pretraining procedure, with tf32 precision, a fused AdamW optimizer,
and \texttt{torch.compile}. Both these settings and the 200{,}000-step
total used for extrapolation are taken from the publicly available
configuration of Chronos T5-base, the closest model for which this
information is documented, since Chronos-2's own pretraining setup is
not disclosed. This single-step cost is reported in
Table~\ref{tab:raw_measures}; multiplied by 200{,}000 steps, it gives
$C_{\text{pr}}=65.17$ gCO$_2$, which should be read as a lower bound: it
assumes a single GPU, whereas real pretraining runs typically use many
GPUs in parallel.

\begin{figure}[h]
  \centering
  \includegraphics[width=0.5\textwidth]{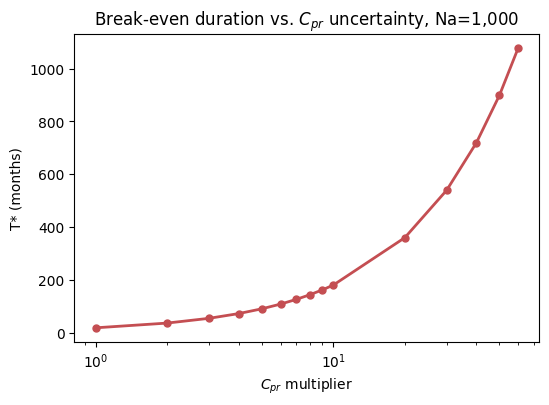}
  \caption{Break-even usage duration $T^*$ between Chronos-2 ZeroShot
  and LightGBM versus a multiplier on $C_{\text{pr}}$, at $N_a=1000$
  (baseline scenario). Beyond $60\times$, no crossover is found within
  our 100-year search horizon.}
  \label{fig:cpr_sensitivity}
\end{figure}

\paragraph{Sensitivity to $C_{\text{pr}}$.}
\label{par:cpr_sensitivity}
$C_{\text{pr}}$ is an extrapolated lower bound: as said, it is estimated on a
single GPU, with a step count borrowed from Chronos T5-base. We test
how the break-even usage duration $T^*$ between Chronos-2 ZeroShot and
LightGBM moves when $C_{\text{pr}}$ is multiplied by a factor $k$ from
$1\times$ to $100\times$, at $N_a=1000$ under the baseline scenario
(Figure~\ref{fig:cpr_sensitivity}). $T^*$ is proportional to
$C_{\text{pr}}$: $T^* \approx 18\,k$ months. $C_{\text{pr}}$ is a fixed
cost repaid by a constant monthly saving of Chronos-2 over LightGBM, so
scaling it scales the time needed to repay it. A $2\times$ higher cost
doubles the break-even to 36 months, $5\times$ gives about 90 months,
and $10\times$ gives 180 months (15 years). Beyond $60\times$
($\approx$90 years), Chronos-2 does not break even within our 100-year
search horizon. A higher pretraining cost lengthens the break-even
rather than shortening it. Since our estimate is a lower bound, the
18-month figure at $N_a=1000$ should be read as optimistic for
Chronos-2, which strengthens the conclusion that a TSFM is not
amortized in every deployment. This analysis varies $C_{\text{pr}}$
only, all other parameters staying at their baseline values.

\paragraph{Estimating training FLOPs from a forward pass.}
For PatchTST and Chronos-2, we measure FLOPs with \texttt{fvcore}, which
traces a model's forward computation graph but not its backward pass,
generated dynamically by autograd. We estimate a full training step as
$3\times$ the forward pass alone, following the scaling law analysis of \citet{kaplan2020scaling}, confirmed empirically across a range of
architectures by ~\citet{epochai2022backward}. LightGBM's FLOPs are computed
analytically from its histogram-based algorithm instead, since
\texttt{fvcore} does not apply to tree-based models.

\paragraph{Dataset and number of runs for Part I.}
All measurements of Part I are carried out on ETTh1, an hourly electricity
transformer temperature dataset with 7 variables, split 70/15/15 into
train, validation, and test sets, yielding 12{,}026 training windows for
LightGBM and PatchTST. \textbf{Every carbon cost figure in the tables below is averaged over
$N=10$ independent runs for training, fine-tuning, and pretraining-step
measurements, and over $N=20$ independent runs for inference
measurements, each inference run itself averaging 1{,}000 individual
calls.}

\paragraph{Fixed training budget.}
Training LightGBM and PatchTST for carbon measurement requires choosing
a training length, since neither has a natural stopping point; we fix
this arbitrarily to 300 estimators and 20 epochs respectively, applied
identically in every run.

\paragraph{Model hyperparameters.}
Table~\ref{tab:hyperparams} reports the hyperparameters used for 
LightGBM, PatchTST, and Chronos-2 fine-tuning/pretraining measurements for the carbon cost model.

\begin{table}[h]
    \centering
    \caption{Hyperparameters used for the models in Section~\ref{sec:cost_model}.}
    \label{tab:hyperparams}
    \small
    \begin{tabular}{lll}
        \toprule
        Model & Parameter & Value \\
        \midrule
        \multirow{6}{*}{LightGBM} & Estimators & 300 \\
                                  & Num.\ leaves & 31 \\
                                  & Max bin & 255 \\
                                  & Learning rate & 0.05 \\
                                  & Feature/bagging fraction & 0.8 \\
                                  & Bagging frequency & 5 \\
        \midrule
        \multirow{6}{*}{PatchTST} & Patch length / stride & 16 / 8 \\
                                  & $d_{\text{model}}$ & 128 \\
                                  & Num.\ heads & 4 \\
                                  & Num.\ layers & 3 \\
                                  & Dropout & 0.2 \\
                                  & Epochs / batch size & 20 / 32 \\
                                  & Learning rate & $10^{-3}$ \\
        \midrule
        \multirow{4}{*}{Chronos-2 (pretraining step)} 
                                  & Context length & 2048 \\
                                  & Batch size & 32 \\
                                  & Learning rate & $10^{-3}$ \\
                                  & Steps per run (extrapolated to) & 100 (200{,}000) \\
        \midrule
        \multirow{3}{*}{Chronos-2 (LoRA fine-tuning)} 
                                  & Steps & 2{,}000 \\
                                  & Batch size & 32 \\
                                  & Learning rate & $10^{-4}$ \\
        \midrule
        \multirow{3}{*}{Chronos-2 (full fine-tuning)} 
                                  & Epochs & 3 \\
                                  & Batch size & 32 \\
                                  & Learning rate & $10^{-5}$ \\
        \bottomrule
    \end{tabular}
\end{table}

\paragraph{Fine-tuning parameters.}
$C_{\text{ft}}^{\text{LoRA}}$ is measured directly at 2{,}000 steps,
matching the value used in Section~\ref{sec:cost_model}. $\beta$ is
reported per training window rather than as a fixed cost, since full
fine-tuning's total cost scales with $N_a$ through $W_{\text{Full}}(N_a)$.

\noindent Two tables report the results of this protocol, and answer
different questions. Table~\ref{tab:raw_measures} gives the cost of
fixed, non-scenario operations, one full training run and one inference
call, on both CPU and GPU where measured, except for PatchTST training
(GPU only, its realistic training device) and Chronos-2 training (GPU
only, since it is never trained from scratch). For Chronos-2, the
training row is the single pretraining step described above, not a full
run. Table~\ref{tab:parameter_values} instead gives the per-unit parameters
plugged into the cost equations of Section~\ref{sec:cost_model}, from a
separate measurement campaign than Table~\ref{tab:raw_measures}: for
LightGBM, on CPU (LightGBM is assumed to run on CPU in production); for
PatchTST and Chronos-2, on GPU, their realistic deployment device.

\begin{table}[h]
  \centering
  \caption{Raw measures: carbon cost, FLOPs, and latency for one full
  training run and one inference call, per approach (mean $\pm$ std).
  Chronos-2's training row is a single pretraining step, not a full run.}
  \label{tab:raw_measures}
  \begin{tabular}{llccc}
    \toprule
    Model & Operation & Carbon (gCO$_2$) & FLOPs & Latency (ms) \\
    \midrule
    \multirow{2}{*}{LightGBM} & Training, CPU & $0.00722 \pm 0.00070$ & \multirow{2}{*}{$5.32 \times 10^{10}$} & $2{,}903.6 \pm 163.7$ \\
                              & Training, GPU & $0.01769 \pm 0.00078$ &                                        & $5{,}698.1 \pm 222.5$ \\
    \cmidrule(l){2-5}
                              & Inference, CPU & $(6.81 \pm 0.39) \times 10^{-6}$ & \multirow{2}{*}{$2{,}972.5$} & $2.17 \pm 0.57$ \\
                              & Inference, GPU & $(6.99 \pm 1.42) \times 10^{-6}$ &                              & $2.33 \pm 1.04$ \\
    \midrule
    PatchTST & Training, GPU  & $1.0800 \pm 0.0087$ & $5.48 \times 10^{14}$ & $185{,}478.9 \pm 784.9$ \\
    \cmidrule(l){2-5}
             & Inference, CPU & $(3.26 \pm 0.15) \times 10^{-5}$ & \multirow{2}{*}{$7.61 \times 10^{8}$} & $8.70 \pm 13.87$ \\
             & Inference, GPU & $(6.60 \pm 0.42) \times 10^{-6}$ &                                        & $2.70 \pm 0.12$ \\
    \midrule
    Chronos-2 & Pretraining, GPU (1 step) & $(3.26 \pm 0.02) \times 10^{-4}$ & $1.45 \times 10^{12}$ & $53.80 \pm 0.41$ \\
    \cmidrule(l){2-5}
              & Inference, CPU & $(3.70 \pm 4.13) \times 10^{-4}$ & \multirow{2}{*}{$1.51 \times 10^{9}$} & $73.37 \pm 53.92$ \\
              & Inference, GPU & $(6.08 \pm 0.05) \times 10^{-5}$ &                                        & $23.64 \pm 0.96$ \\
    \bottomrule
  \end{tabular}
\end{table}

\begin{table}[h]
  \centering
  \caption{Cost model parameters (Section~\ref{sec:cost_model}).}
  \label{tab:parameter_values}
  \begin{tabular}{lc}
    \toprule
    Parameter & Value \\
    \midrule
    $\alpha$ (gCO$_2$/window)              & $5.51 \times 10^{-7}$ \\
    $C_{\text{inf}}^{\text{C}}$ (gCO$_2$)  & $6.44 \times 10^{-6}$ \\
    $\alpha_{\text{ptst}}$ (gCO$_2$/window)& $9.01 \times 10^{-5}$ \\
    $C_{\text{inf}}^{\text{ptst}}$ (gCO$_2$) & $7.43 \times 10^{-6}$ \\
    $C_{\text{pr}}$ (gCO$_2$)              & $65.17$ \\
    $C_{\text{inf}}^{\text{T}}$ (gCO$_2$)  & $6.44 \times 10^{-5}$ \\
    $C_{\text{ft}}^{\text{LoRA}}$ (gCO$_2$)& $0.62$ \\
    $\beta$ (gCO$_2$/window)               & $2.21 \times 10^{-5}$ \\
    \bottomrule
  \end{tabular}
\end{table}

\paragraph{Cost measurements for the pruning experiments.}
\label{sec:appendix_protocol_pruning}

Full and pruned models are measured under identical conditions, on the
hardware and with the CodeCarbon configuration described above. CO$_2$
emissions are accumulated over zero-shot inference on a fixed subset of
the calibration tasks (\emph{ercot}, \emph{exchange\_rate},
\emph{monash\_hospital}, \emph{monash\_m3\_monthly}, \emph{m4\_yearly},
and a fixed random sample of 200 series of \emph{dominick}; 25,411
windows in total), identical for both models, rather than on the full
25-task set, to keep repeated measurements tractable. We report the
mean and standard deviation over 5 repetitions of this whole pass.
Latency is measured on 10 series drawn at random (fixed seed) from the
same subset, with 3 untimed warmup passes and 20 timed runs per series;
we report the median and standard deviation over the 200 timed calls.
FLOPs are counted with \texttt{fvcore} for a single forward pass of the
model on a random input of 8192 points (the model's context length) with
batch size 1. They depend on the input shape and not on its values, so
one deterministic count per model suffices. Both models run on the same
grid, so the relative CO$_2$ reduction does not depend on the carbon
intensity applied.

\newpage
\section{Appendix 3: Cost Model Results}
\subsection{Figures}
\label{sec:appendix_figures}

\begin{figure}[h]
  \centering
  \includegraphics[width=0.9\textwidth]{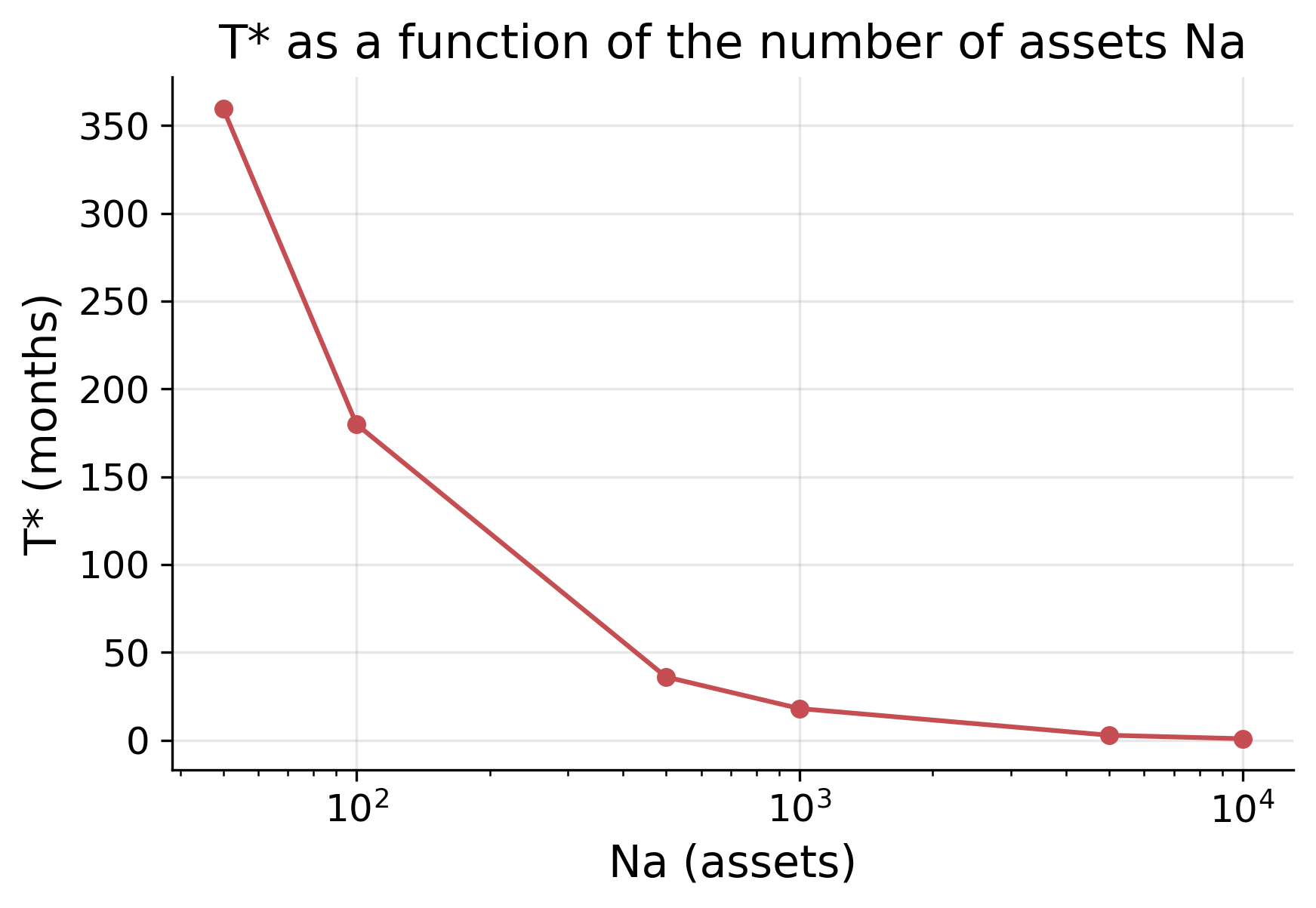}
  \caption{Break-even usage duration $T^*$ (months) as a function of the number of assets $N_a$, at which Chronos-2 ZeroShot becomes cheaper than LightGBM under the baseline scenario.}
  \label{fig:new/fig2_t_star_vs_na.png}
\end{figure}

\newpage

\section{Appendix 4: Model Pruning}
\label{sec:appendix_pruning}

\paragraph{Clarification and additional ablation of the pruning criterion}

To clarify, the block removed in our main experiments
(\S\ref{sec:setuppruning}) is block 6, the block with the \emph{highest}
cosine score (0.963) among the 12 encoder blocks, identified by our
criterion as the most redundant. Block 11, which has the \emph{lowest}
cosine score (0.576), is the block the criterion identifies as the
\emph{most} important to keep, and is preserved rather than removed in
all configurations we evaluate. Table~\ref{tab:all_block_scores}
reports the score for every block.

To verify that our criterion correctly identifies block 11 as critical
rather than redundant, we removed it on its own and evaluated the
resulting model on the 6 calibration tasks, checking whether its low
cosine score does correspond to a substantial performance drop. As an additional check, we removed one arbitrarily chosen non-extreme
block not selected by our criterion (block 5), which our criterion
also happens to identify as redundant and ranks close to, though
slightly lower than, block 6 (cosine score 0.954 versus 0.963), and
evaluated the resulting model on the same 6 calibration tasks.

Table~\ref{tab:block_ablation} reports the aggregated skill scores
obtained. Removing block 11 causes a severe degradation (MASE skill
score dropping from 9.57\% to $-$47.30\%, WQL from 26.89\% to
$-$63.09\%), confirming that its low cosine score correctly identifies
it as critical to keep. By contrast, removing block 5 or block 6, both
identified as redundant by the criterion, has a comparably modest
effect on performance, consistent with their similarly high cosine
scores.

\begin{table}[h]
    \centering
    \caption{Aggregated MASE and WQL skill scores on the 6 calibration
    tasks, removing a single block at a time.}
    \label{tab:block_ablation}
    \begin{tabular}{lccc}
        \toprule
        Removed block & Cosine score & MASE skill (\%) & WQL skill (\%) \\
        \midrule
        None (full model)  & --    & 9.57   & 26.89 \\
        Block 6 (retained) & 0.963 & 10.24  & 28.65 \\
        Block 5            & 0.954 & 9.16   & 28.40 \\
        Block 11           & 0.576 & $-$47.30 & $-$63.09 \\
        \bottomrule
    \end{tabular}
\end{table}

\begin{table}[h]
    \centering
    \caption{Mean cosine score per encoder block, averaged across the 6
    calibration tasks (median within task, equal weight across tasks).
    Lower scores indicate less redundant blocks.}
    \label{tab:all_block_scores}
    \begin{tabular}{cc}
        \toprule
        Block & Cosine score \\
        \midrule
        6  & 0.963 \\
        4  & 0.960 \\
        2  & 0.957 \\
        3  & 0.955 \\
        5  & 0.954 \\
        1  & 0.946 \\
        7  & 0.936 \\
        8  & 0.936 \\
        9  & 0.906 \\
        10 & 0.871 \\
        0  & 0.852 \\
        11 & 0.576 \\
        \bottomrule
    \end{tabular}
\end{table}

Table~\ref{tab:benchmark2_stats} reports the full task-level statistics for the 27 tasks of Chronos Benchmark~II used in our evaluation (25 tasks after excluding the two ETT variants, see Section~\ref{sec:setuppruning}). Table~\ref{tab:per_task_results} reports the per-task MASE and WQL scores for the full and pruned models, complementing the aggregated results of Section~\ref{sec:results}.

\begin{table}[h]
    \centering
    \caption{Task-level statistics for the 27 tasks of Chronos
    Benchmark~II, reproduced from \citet{ansari2024chronos} (Table~3).
    \emph{ETT (15~Min.)} and \emph{ETT (Hourly)} are excluded from our evaluation due to a data loading
    incompatibility; our study covers the
    remaining 25 tasks.}
    \label{tab:benchmark2_stats}
    \small
    \begin{tabular}{llcccccc}
        \toprule
        Dataset & Domain & Freq. & \# Series & \multicolumn{3}{c}{Series length} & $H$ \\
        \cmidrule(lr){5-7}
        & & & & min & avg & max & \\
        \midrule
        Australian Electricity & energy      & 30min & 5     & 230736 & 231052 & 232272 & 48 \\
        CIF 2016                & banking     & 1M    & 72    & 28     & 98     & 120    & 12 \\
        Car Parts                & retail      & 1M    & 2674  & 51     & 51     & 51     & 12 \\
        Covid Deaths              & healthcare  & 1D    & 266   & 212    & 212    & 212    & 30 \\
        Dominick                  & retail      & 1D    & 100014 & 201   & 296    & 399    & 8  \\
        ERCOT Load                 & energy      & 1H    & 8     & 154854 & 154854 & 154854 & 24 \\
        ETT (15~Min.)    & energy      & 15min & 14    & 69680  & 69680  & 69680  & 24 \\
        ETT (Hourly)      & energy      & 1H    & 14    & 17420  & 17420  & 17420  & 24 \\
        Exchange Rate               & finance     & 1B    & 8     & 7588   & 7588   & 7588   & 30 \\
        FRED-MD                      & economics   & 1M    & 107   & 728    & 728    & 728    & 12 \\
        Hospital                      & healthcare  & 1M    & 767   & 84     & 84     & 84     & 12 \\
        M1 (Monthly)                   & various     & 1M    & 617   & 48     & 90     & 150    & 18 \\
        M1 (Quarterly)                  & various     & 3M    & 203   & 18     & 48     & 114    & 8  \\
        M1 (Yearly)                      & various     & 1Y    & 181   & 15     & 24     & 58     & 6  \\
        M3 (Monthly)                      & various     & 1M    & 1428  & 66     & 117    & 144    & 18 \\
        M3 (Quarterly)                     & various     & 3M    & 756   & 24     & 48     & 72     & 8  \\
        M3 (Yearly)                         & various     & 1Y    & 645   & 20     & 28     & 47     & 6  \\
        M4 (Quarterly)                       & various     & 3M    & 24000 & 24     & 100    & 874    & 8  \\
        M4 (Yearly)                           & various     & 1Y    & 23000 & 19     & 37     & 841    & 6  \\
        M5                                     & retail      & 1D    & 30490 & 124    & 1562   & 1969   & 28 \\
        NN5 (Daily)                             & finance     & 1D    & 111   & 791    & 791    & 791    & 56 \\
        NN5 (Weekly)                             & finance     & 1W    & 111   & 113    & 113    & 113    & 8  \\
        Tourism (Monthly)                         & various     & 1M    & 366   & 91     & 298    & 333    & 24 \\
        Tourism (Quarterly)                        & various     & 1Q    & 427   & 30     & 99     & 130    & 8  \\
        Tourism (Yearly)                            & various     & 1Y    & 518   & 11     & 24     & 47     & 4  \\
        Traffic                                      & transport   & 1H    & 862   & 17544  & 17544  & 17544  & 24 \\
        Weather                                       & nature      & 1D    & 3010  & 1332   & 14296  & 65981  & 30 \\
        \bottomrule
    \end{tabular}
\end{table}

\begin{table}[h]
    \centering
    \caption{Per-task comparison between the full and pruned Chronos-2 models
on the 25 tasks of Chronos Benchmark~II. Lower MASE and WQL indicate
better performance. Bold values mark the better-performing configuration
on each task, based on full-precision values even where rounded figures
appear equal.}
    \label{tab:per_task_results}
    \small
    \begin{tabular}{lcccccc}
        \toprule
        Task & \multicolumn{2}{c}{MASE} & \multicolumn{2}{c}{WQL} \\
        \cmidrule(lr){2-3} \cmidrule(lr){4-5}
        & Full & Pruned & Full & Pruned \\
        \midrule
        dominick & 0.809 & \textbf{0.796} & 0.324 & \textbf{0.321} \\
        ercot & 0.832 & \textbf{0.822} & 0.026 & \textbf{0.026} \\
        exchange\_rate & 1.852 & \textbf{1.807} & 0.012 & \textbf{0.011} \\
        m4\_quarterly & 1.201 & \textbf{1.186} & 0.076 & \textbf{0.075} \\
        m4\_yearly & \textbf{3.378} & 3.427 & \textbf{0.117} & 0.118 \\
        m5 & \textbf{0.905} & 0.907 & \textbf{0.553} & 0.553 \\
        monash\_australian\_electricity & \textbf{0.629} & 0.744 & \textbf{0.031} & 0.036 \\
        monash\_car\_parts & 0.834 & \textbf{0.834} & 0.961 & \textbf{0.958} \\
        monash\_cif\_2016 & 0.933 & \textbf{0.898} & 0.011 & \textbf{0.010} \\
        monash\_covid\_deaths & 35.406 & \textbf{32.476} & 0.039 & \textbf{0.029} \\
        monash\_fred\_md & \textbf{0.469} & 0.592 & \textbf{0.020} & 0.024 \\
        monash\_hospital & 0.773 & \textbf{0.766} & 0.054 & \textbf{0.052} \\
        monash\_m1\_monthly & \textbf{1.027} & 1.037 & 0.147 & \textbf{0.146} \\
        monash\_m1\_quarterly & \textbf{1.661} & 1.700 & \textbf{0.083} & 0.085 \\
        monash\_m1\_yearly & 3.830 & \textbf{3.825} & 0.138 & \textbf{0.133} \\
        monash\_m3\_monthly & \textbf{0.813} & 0.815 & \textbf{0.089} & 0.089 \\
        monash\_m3\_quarterly & \textbf{1.162} & 1.213 & \textbf{0.070} & 0.073 \\
        monash\_m3\_yearly & \textbf{2.996} & 3.084 & \textbf{0.147} & 0.150 \\
        monash\_nn5\_weekly & \textbf{0.882} & 0.891 & \textbf{0.082} & 0.082 \\
        monash\_tourism\_monthly & \textbf{1.423} & 1.436 & \textbf{0.073} & 0.077 \\
        monash\_tourism\_quarterly & \textbf{1.566} & 1.586 & 0.065 & \textbf{0.063} \\
        monash\_tourism\_yearly & \textbf{3.666} & 3.780 & \textbf{0.152} & 0.154 \\
        monash\_traffic & \textbf{0.818} & 0.819 & 0.242 & \textbf{0.237} \\
        monash\_weather & \textbf{0.774} & 0.777 & \textbf{0.125} & 0.126 \\
        nn5 & 0.577 & \textbf{0.559} & 0.149 & \textbf{0.144} \\
        \bottomrule
    \end{tabular}
\end{table}
\clearpage
\section{Appendix 5: Limitations}
\label{sec:limitations}

\subsection{Carbon cost modeling}
\label{app:limitations_carbon}

\paragraph{Scope of the measurements.}
All carbon measurements are specific to a single hardware setup (NVIDIA
A100 80GB, AMD EPYC 7V13), a single dataset (ETTh1), and a single grid
carbon intensity ($\approx$56 gCO$_2$/kWh, Île-de-France). Absolute
values would differ on other hardware, datasets, or regions, and our
results should be read as illustrative of a scenario rather than
universal constants.

\paragraph{CPU power estimation.}
CPU power is estimated from a TDP-based model rather than measured
directly, since the RAPL hardware counters CodeCarbon otherwise reads
were not accessible on our virtual machine. This affects every
measurement using this fallback, but disproportionately LightGBM, which
we assume runs on CPU in production. 

\paragraph{Estimating $C_{\text{pr}}$.}
$C_{\text{pr}}$, the parameter driving every break-even result in this
paper, is extrapolated from a single measured pretraining step using
the publicly documented step count of Chronos T5-base, since Chronos-2's
own pretraining schedule and infrastructure are not disclosed. It should
be read as a lower bound: it assumes a single GPU rather than the
distributed setup a real pretraining run would likely use.


\paragraph{Comparison at unequal footing.}
Each model is costed under the usage pattern it was built for
(Section~\ref{sec:cost_model}): LightGBM and PatchTST are trained per
asset, while a single LoRA adapter is shared across assets for
Chronos-2. Sharing a single PatchTST across similar assets would remove
its dependence on $N_a$ in training cost and would likely make it much
cheaper than reported here. The comparison is thus not at equal
informational access, and PatchTST's cost should be read as that of
per-asset training. A matched comparison requires a separate
measurement campaign.

\paragraph{Hyperparameter tuning cost.}
The training budgets of LightGBM (300 estimators) and PatchTST (20
epochs) are fixed, untuned choices, and no hyperparameter search was
run, so no tuning cost enters the cumulative cost of these baselines.
In practice, a search over $K$ configurations would add roughly $K$
training runs per tuned model. Chronos-2 zero-shot requires none, so
omitting it favors the baselines: including it would move the
break-even earlier. Also, since LightGBM's cost grows roughly linearly with the number of
estimators, a smaller model would shift the break-even later, and a
larger one earlier. We did not quantify this effect. 

\paragraph{Consistency across measurement campaigns.}
The parameters used in the cost equations (Table~\ref{tab:parameter_values})
come from a separate, earlier measurement campaign than the raw measures
reported in Table~\ref{tab:raw_measures}.

\paragraph{Measurement variance.}
One measurement, Chronos-2's CPU inference cost, shows unusually high
variance across runs (Table~\ref{tab:raw_measures}) that we did not
fully investigate. We report it as measured rather than excluding it,
but it should be treated with more caution than the rest of the table.

\subsection{Pruning}
\label{app:limitations_pruning}

\paragraph{Benchmark coverage.}
Two tasks of Chronos Benchmark~II (\emph{ETT 15-min} and \emph{ETT
hourly}) are excluded from our evaluation due to a technical
incompatibility between their loading mechanism and the current version
of the HuggingFace \texttt{datasets} library; our results cover 25 of
the 27 original tasks.

\paragraph{Pruning granularity.}
Pruning operates at the level of entire encoder blocks. Removing a
block removes its time attention, group attention, and feed-forward
sub-layers simultaneously, so we cannot isolate which specific
mechanism drives the observed redundancy or performance impact.

\paragraph{Block selection at fixed compression.}
At each compression level $k$, we only evaluate the specific set of $k$
blocks chosen by the cosine ordering. We do not compare this selection
against alternative sets of $k$ blocks chosen by other criteria, \textbf{so we
cannot claim that the cosine score identifies the best possible blocks
to remove at a given compression budget, only that the resulting
configuration performs well in absolute terms}.

\paragraph{Univariate evaluation only.}
All experiments are run in strict univariate mode
(\texttt{cross\_learning=False}). We do not evaluate whether pruning
affects Chronos-2's in-context learning capabilities across series,
which depend on batch composition and are only engaged in
cross-learning mode.

\paragraph{Fixed compute-cost measurement subset.}
CO$_2$ and latency are measured on a fixed subset of tasks (25,411
windows total) rather than the full 25-task evaluation set, to keep
repeated measurements tractable. While inference cost primarily depends
on input size rather than task content, this subset is not identical to
the one used for the performance evaluation in
Table~\ref{tab:full_vs_pruned}.

\paragraph{Single hardware and single run.}
All compute cost measurements are collected on a single machine (see
Appendix~\ref{sec:appendix_protocol}) and, for the main performance
comparison (Table~\ref{tab:full_vs_pruned}), from a single deterministic
zero-shot evaluation run per configuration.

\paragraph{Scope of the pruning criterion.}
The cosine similarity score is a lightweight, gradient-free heuristic.
We do not compare it against alternative pruning criteria (e.g.,
gradient-based importance scores), so we cannot claim it is the most
effective method available, only that it is computationally cheap and
yields a configuration with no measurable aggregated performance loss on
this benchmark.

\clearpage

\section*{NeurIPS Paper Checklist}



\begin{enumerate}

\item {\bf Claims}
    \item[] Question: Do the main claims made in the abstract and introduction accurately reflect the paper's contributions and scope?
    \item[] Answer: \answerYes{} 
    \item[] Justification: Yes

\item {\bf Limitations}
    \item[] Question: Does the paper discuss the limitations of the work performed by the authors?
    \item[] Answer: \answerYes{} 
    \item[] Justification: Yes, see Appendix \ref{sec:limitations}

\item {\bf Theory assumptions and proofs}
    \item[] Question: For each theoretical result, does the paper provide the full set of assumptions and a complete (and correct) proof?
    \item[] Answer: \answerNA{}.
    \item[] Justification: The paper doesn't include any theoretical result.

    \item {\bf Experimental result reproducibility}
    \item[] Question: Does the paper fully disclose all the information needed to reproduce the main experimental results of the paper to the extent that it affects the main claims and/or conclusions of the paper (regardless of whether the code and data are provided or not)?
    \item[] Answer: \answerYes{}
    \item[] Justification: The Appendix \ref{sec:appendix_protocol} explains all the details to reproduce the carbon cost modelization, and Section \ref{sec:setuppruning} details the setup to reproduce the pruning experiences.

\item {\bf Open access to data and code}
    \item[] Question: Does the paper provide open access to the data and code, with sufficient instructions to faithfully reproduce the main experimental results, as described in supplemental material?
    \item[] Answer: \answerNo{}
    \item[] Justification: While the code is not available, we describe all the steps to reproduce it. Data is accessible.

\item {\bf Experimental setting/details}
    \item[] Question: Does the paper specify all the training and test details (e.g., data splits, hyperparameters, how they were chosen, type of optimizer) necessary to understand the results?
    \item[] Answer: \answerYes{}
    \item[] Justification: Yes (see answer 4)

\item {\bf Experiment statistical significance}
    \item[] Question: Does the paper report error bars suitably and correctly defined or other appropriate information about the statistical significance of the experiments?
    \item[] Answer: \answerNo{}
    \item[] Justification: In the first part : partially. Raw carbon cost, FLOPs, and latency measurements are reported with standard deviations over multiple independent runs (Table~\ref{tab:raw_measures}, Appendix~\ref{sec:appendix_protocol}). The cost model figures (Figures~\ref{fig:new/fig1_carbon_baseline}, \ref{fig:fig3_drift_sensitivity}, \ref{fig:fig4_frequency_sensitivity}) show simulated cumulative cost curves without uncertainty bands, for visual clarity. In the second part, our goal is to assess whether pruning can reduce Chronos-2's compute cost without substantially degrading forecasting performance, rather than to establish statistical superiority of one configuration over another. We report point estimates (aggregated skill scores, win rate) for this comparison across the 25 evaluation tasks, without a formal significance test on this specific result. We do report standard deviations for CO2 emissions and latency, and a Wilcoxon signed-rank test for the block-11 redundancy analysis.

\item {\bf Experiments compute resources}
    \item[] Question: For each experiment, does the paper provide sufficient information on the computer resources (type of compute workers, memory, time of execution) needed to reproduce the experiments?
    \item[] Answer: \answerYes{}
    \item[] Justification: Yes, see Appendix \ref{sec:appendix_protocol}

\item {\bf Code of ethics}
    \item[] Question: Does the research conducted in the paper conform, in every respect, with the NeurIPS Code of Ethics \url{https://neurips.cc/public/EthicsGuidelines}?
    \item[] Answer: \answerYes{}
    \item[] Justification: Yes

\item {\bf Broader impacts}
    \item[] Question: Does the paper discuss both potential positive societal impacts and negative societal impacts of the work performed?
    \item[] Answer: \answerYes{}
    \item[] Justification: This paper directly addresses the environmental cost of machine learning deployment. It provides a practical framework for reducing CO$_2$ emissions associated with TSFM deployment, and demonstrates that structured pruning can reduce inference-time emissions by 6 to 9\% at no cost to forecasting performance.

\item {\bf Safeguards}
    \item[] Question: Does the paper describe safeguards that have been put in place for responsible release of data or models that have a high risk for misuse (e.g., pre-trained language models, image generators, or scraped datasets)?
    \item[] Answer: \answerNA{}
    \item[] Justification: the paper poses no such risks.

\item {\bf Licenses for existing assets}
    \item[] Question: Are the creators or original owners of assets (e.g., code, data, models), used in the paper, properly credited and are the license and terms of use explicitly mentioned and properly respected?
    \item[] Answer: \answerYes{}
    \item[] Justification: Yes

\item {\bf New assets}
    \item[] Question: Are new assets introduced in the paper well documented and is the documentation provided alongside the assets?
    \item[] Answer: \answerNA{}
    \item[] Justification: the paper does not release new assets.

\item {\bf Crowdsourcing and research with human subjects}
    \item[] Question: For crowdsourcing experiments and research with human subjects, does the paper include the full text of instructions given to participants and screenshots, if applicable, as well as details about compensation (if any)?
    \item[] Answer: \answerNA{}
    \item[] Justification: the paper does not involve crowdsourcing nor research with human subjects.

\item {\bf Institutional review board (IRB) approvals or equivalent for research with human subjects}
    \item[] Question: Does the paper describe potential risks incurred by study participants, whether such risks were disclosed to the subjects, and whether Institutional Review Board (IRB) approvals (or an equivalent approval/review based on the requirements of your country or institution) were obtained?
    \item[] Answer: \answerNA{}
    \item[] Justification: the paper does not involve crowdsourcing nor research with human subjects.

\item {\bf Declaration of LLM usage}
    \item[] Question: Does the paper describe the usage of LLMs if it is an important, original, or non-standard component of the core methods in this research? Note that if the LLM is used only for writing, editing, or formatting purposes and does \emph{not} impact the core methodology, scientific rigor, or originality of the research, declaration is not required.
    \item[] Answer: \answerNA{}
    \item[] Justification: only for writing, editing, or formatting purposes

\end{enumerate}

\end{document}